%% file: main.tex
\documentclass[10pt]{article}

\usepackage[utf8]{inputenc} 
\usepackage[T1]{fontenc}    

\usepackage{amsfonts}
\usepackage{amsmath}
\usepackage{amscd}
\usepackage{amsthm}
\usepackage{amssymb}
\usepackage{amsfonts}       
\usepackage{algorithm}
\usepackage{algpseudocode}
\usepackage{booktabs}       

\usepackage{url}            

\usepackage{nicefrac}       
\usepackage{microtype}      
\usepackage{xcolor}         
\usepackage{natbib}
\usepackage[colorlinks,linkcolor=blue,citecolor=blue]{hyperref}
\usepackage{helvet}  
\usepackage{courier}  
\usepackage{verbatim}
\usepackage{graphicx}  
\usepackage{xcolor}

\usepackage{enumitem}
\usepackage{pifont}
\usepackage{stmaryrd}
\usepackage{tabularx}
\usepackage{makecell}
\usepackage{array}
\usepackage{bbm}
\usepackage{multirow}

\usepackage[verbose=true,letterpaper]{geometry}
\newgeometry{
    textheight=9in,
    textwidth=6in,
    top=1in,
    headheight=12pt,
    headsep=25pt,
    footskip=30pt
}

\input{custom_command.tex}

\begin{document}
\title{\textbf{On the Discussion of Large Language Models:}\\\textbf{ Symmetry of Agents and Interplay with Prompts}}
\author{Qineng Wang$^{*\dag}$, Zihao Wang$^{*\ddag}$, Ying Su$^\ddag$, Yangqiu Song$^\ddag$\\\\
$\dag$  {\normalsize College of Computer Science and Technology, Zhejiang University;} \\{\normalsize\texttt{wongqineng73@zju.edu.cn}}\\
$\ddag$ {\normalsize CSE, HKUST; \texttt{\{zwanggc,ysuay,yqsong\}@cse.ust.hk}}
}
\date{}
\maketitle
\def\thefootnote{*}\footnotetext{These authors contributed equally to this work}\def\thefootnote{\arabic{footnote}}
\begin{abstract}
Two ways has been discussed to unlock the reasoning capability of a large language model. The first one is prompt engineering and the second one is to combine the multiple inferences of large language models, or the multi-agent discussion. Theoretically, this paper justifies the multi-agent discussion mechanisms from the symmetry of agents. Empirically, this paper reports the empirical results of the interplay of prompts and discussion mechanisms, revealing the empirical state-of-the-art performance of complex multi-agent mechanisms can be approached by carefully developed prompt engineering. This paper also proposes a scalable discussion mechanism based on conquer and merge, providing a simple multi-agent discussion solution with simple prompts but state-of-the-art performance.
\end{abstract}

\input{content_arxiv.tex}

\clearpage
\bibliographystyle{plainnat}
\bibliography{ref}
\clearpage
\input{appendix.tex}

\end{document}

%% file: custom_command.tex
\theoremstyle{plain}

\newtheorem{definition}{Definition}

\newcommand{\llm}{\mathcal{L}}
\newcommand{\mech}{\mathcal{M}}
\newcommand{\task}{\mathcal{T}}

\newcommand{\cmark}{\ding{51}}
\newcommand{\xmark}{\ding{55}}
\newcommand{\boxcheck}{\makebox[1.5em][l]{\makebox[2pt][l]{{\Large$\boxempty$}}{\Large\cmark}}}
\newcommand{\boxwrong}{\makebox[1.5em][l]{\makebox[2pt][l]{{\Large$\boxempty$}}{\Large\xmark}}}

%% file: content_arxiv.tex

\section{Introduction}

Let $\llm$ be a Large Language Model (LLM) and $x$ be the input text, the forward inference generates a response of surprising quality, which is written as
\begin{align}
    \hat{y} = \llm(x).
\end{align}

Given the high price of obtaining one large language model, one essential research question is to \textit{unlock the reasoning capability} of large language models so that they can perform better on various tasks. One of the key directions is prompt engineering, where the format and content of the input texts are decorated to improve performances for downstream tasks. Specifically, for a task $\task$, the original input $x$ is decorated as a task-specific prompt input $p(x;\task,\llm)$. The function $p(\cdot; \task, \llm)$ is denoted as a prompt decorator for $\task$ and $\llm$. Then, output $\hat y$ is generated by prompt engineering, which is written as
\begin{align}\label{eq:prompt-engineering}
    \hat{y} = \llm(p(x; \task, \llm)).
\end{align}
A seminal work of design the \textit{prompt decorator} is Chain-of-Thought (CoT), demonstrating that simple modifications of the original input make a significant difference~\citep{cot}. Other effective approaches that make more powerful $p(\cdot, \task)$ include as In-Context Learning (ICL)~\citep{gpt3} and emotional prompt~\citep{emoprompt}. 

Another way to improve the reasoning capability of $\llm(x)$ is to leverage multiple times of inference of large language models, leading to a way larger space for \textbf{mechanism design}. The output $\hat{y}$ derived by mechanism $\mech$ is considered as
\begin{align}\label{eq:overall-framework}
    \hat{y} = M\left(x; \{ (\llm_i, p_i(\cdot, \task, \llm_i))\}_{i=1}^n\right),
\end{align}
where $M$ is the mechanism pipeline, $\llm_i$ is the $i$-th inference of LLM, and $p_i(\cdot, \task, \llm_i)$ is the corresponding prompt decorator. A mechanism $\mech$ is designed by jointly considering $\mech=(M, \{(\llm_i, p_i)\}_{i=1}^n)$, i.e., how to organize prompt decorated LLM inferences into the mechanism pipeline $M$.
Some representative designs of the mechanism pipeline $M$ include fixed pipelines, e.g. Chain-of-Thoughts Self-Consistency (CoT-SC)~\citep{cot-sc}, or adaptive pipelines by heuristic designs, such as Tree-of-Thoughts (ToT)~\citep{tot}, Graph-of-Thoughts (GoT)~\citep{got}, and Cumulative Reasoning (CR)~\citep{cr}.

However, it should be stressed that the design of mechanism pipeline $M$ does not suffice to describe the design space of mechanism $\mech$. The underlying reason is that the input and output of large language models are natural languages. The mechanism pipeline $M$ only defines the flow of information explicitly. Meanwhile, what is ignored by $M$ is the fact that texts generated by LLMs can yield \textbf{self-organized} reasoning process and form \textbf{emergent pipelines} of information processing. For example, LLM agents in the debate framework~\citep{debate} affect each other through texts and there is no predefined reasoning process to derive the results. Recently, \citet{mad,reconcile} also devised weak forms of mechanism pipeline $M$ so that the emergent and implicit pipelines rise freely from the process of multi-agent debate. For convenience, the development of the mechanisms $\mech$ above is termed as ``Discussion Engineering'', to emphasize the \textbf{emergent pipelines} introduced by the \textbf{self-organized information processing}.

In this paper, we provide a systematic characterization of the mechanism designs with particular emphasis on the symmetries of agents.
Breaking down the entire mechanism $\mech$ into the inference calls of LLMs as Equation~\eqref{eq:overall-framework} suffices to describe the computation process. Still, this description ignores the conversational behavior that appeared in the multi-agent system, i.e., the affinity between agents and conversation history. In Section~\ref{sec: 3sym}, this affinity is described by the agent-inference assignment which also enables us to understand the computation process by the symmetries of agents.
The existing approaches are then justified by whether the symmetries are broken explicitly. The language of symmetry also provides a new perspective of understanding existing results and our empirical findings.

The second contribution of this paper is the in-depth evaluation of the different combinations of existing mechanism pipelines $M$ and prompt decorators $p_i$. Our evaluation in a refined version of FOLIO dataset \citep{folio}, \textit{FOLIO wiki curated} \citep{cr}, reveals the fact that the gains obtained by designing pipeline $M$ gradually diminish when the prompt decorator gets stronger. Strikingly, the combination of recently proposed mechanisms $M$ with carefully designed prompt decorator $p_i$, as indicated in Equation~\eqref{eq:overall-framework}, performs on par with directly using a good prompt decorator $p$ in Equation~\eqref{eq:prompt-engineering} where $M$ is trivial. This sharp contrast suggests the inherent upper bound of reasoning capability under the \textit{entanglement} of discussion engineering and prompt engineering (Equation~\eqref{eq:overall-framework}) can also be possibly achieved by applying \textbf{strong prompt engineering and weak discussion engineering} (Equation~\eqref{eq:prompt-engineering}).

The third contribution of this paper is also related to the findings above. After revealing both strong prompts and strong mechanisms can approach the inherent upper bound. We explore whether this upper bound can be achieved by simple mechanisms (with more symmetries) instead of explicitly designing asymmetric mechanisms as in previous practices. This idea motivates us to study the behavior of symmetric discussion, whose complexity can be increased by just increasing the number of LLM agents. To reduce the huge cost of organizing a discussion with a large number of LLM agents, we further introduce a scalable mechanism using the concept of group discussion. Our method is called Conquer-and-Merge Discussion (CMD), where the topics are separately discussed in groups and merged with an additional LLM secretary. Empirical evaluation shows that CMD can also approximate the upper bound of the performance by using a simpler prompt decorator. It showcases another pathway that achieves state-of-the-art performance with simple prompt decorators and a simple discussion mechanism.

\section{Related Work}

\subsection{Large language models $\llm$}
With the advent of large language models (LLMs) like GPT-3 \citep{gpt3}, chatGPT \citep{chatgpt}, GPT-4 \citep{gpt4}, Llama series \citep{llama, llama2} and PaLM 2 \citep{palm2}, prompt engineering is a burgeoning field that focuses on optimizing the output of language models by crafting effective input prompts.

\subsection{Prompt decorator $p(\cdot; \task, \llm)$ for reasoning}
Numerous advancements have been made in the field of improving reasoning abilities of LLMs with prompt engineering. Chain of Thought (CoT) \citep{cot, cot0} is a linear problem-solving approach where each step builds upon the previous one. \citet{firstms} propose to apply CoT to multi-step reasoning tasks. To automate the CoT, Auto-CoT \citep{autop} constructs demonstrations by sampling diverse questions and generating reasoning chains. Active-Prompt \citep{activep} aims to select the most uncertain questions for task-specific annotations. Other prompt strategies include PS prompt \citep{pas} which divides task into subtasks then solve them, effective GoT \citep{eGoT} which models human thought processes as a graph rather than a chain, Natural Program \citep{np} which helps to improve the deductive reasoning tasks, re-reading prompt \citep{reread} which revisits the question information embedded within input prompts.

\subsection{Mechanism $\mech$ for reasoning}\label{sec:related-mechanism}
In addition to focusing on the design of prompts themselves, some works incorporate different mechanisms to assist language models in reasoning. 

The design directions of these mechanisms include: 
1. Breaking down the reasoning process into multiple stages, with possible verification at each stage. 
2. Optimizing the reasoning process via ensemble methods.
3. Iteratively prompting the model for reflection and correction. 
4. Utilizing external tools to aid in reasoning. These approaches aim to enhance the overall reasoning capabilities of language models by introducing additional support and guidance throughout the process.\par
\noindent\textbf{Break Down the Reasoning Process into Multiple Stages.}
Self-Polish \citep{sp} make models to progressively refine given problems with multiple stages of prompts. Additionally, some works \citep{decomp, mes, succ, cr} study over decomposing the tasks into smaller tasks, and use an individual prompt to solve each subtask.\par
\noindent\textbf{Optimize the Reasoning Process via Ensemble Methods.}
Tree of Thoughts (ToT) \citep{tot} expands on the reasoning process by considering multiple paths of reasoning and self-evaluating choices. Graph of Thoughts (GoT) \citep{got} further advances this by modeling information as an arbitrary graph, enabling complex networks of thoughts. And some works propose to first sample reasoning pathsvthen vote for the best one, including self-consistency \citep{cot-sc} and step-aware verifier \citep{step-verifier}. Discussion Engineering is also one category of ensemble methods. We will leave it to section \ref{sec:de} for further introduction.\par
\noindent\textbf{Iteratively Prompt the Model for Reflection and Correction.} These works \citep{star, sv, reflexion, sr, sd, rcot} are mainly based on iteratively asking model to find the mistakes or inconsistencies among previous reasoning steps or the knowledge, then solve them one by one. \par
\noindent\textbf{Utilize External Tools or Knowledge to Aid in Reasoning.} Mind's Eye \citep{mindeye} proposes to use physical simulator to help models reason in the physical world. There are also other works \cite{chameleon, mathprompter} will equip models with tools to solve problems. Further more, some works like MemPrompt \citep{memprompt} and LogicSolver \cite{logicsolver} use external knowledge to assist LLMs with reasoning.\par

\subsection{Discussion Engineering - A Subset of Mechanism $\mech$} \label{sec:de}
Self-organized discussion involves multi-agent interactions, where agents spontaneously navigate the entire discussion process and reach a conclusion based on their assigned roles, discussion topics, and framework. Discussion Engineering, a subset of mechanism design $\mech$, focuses on defining crucial components of the discussion framework in self-organized discussions. This includes the discussion process algorithm, roles, utilized Language Models (LLMs), and prompt decorators $p(\cdot; \task, \llm)$ employed at different discussion stages. By leveraging the input interface of an LLM, Discussion Engineering simulates the information exchange between agents and generates corresponding outputs to mimic their speech during ongoing discussions, enabling downstream task-oriented mechanisms.

\noindent\textbf{Recent Works.} Recent progress has been made in the design of Discussion Engineering. The Multi-Agent Debate framework, introduced by \cite{debate}, establishes a mechanism for symmetric discussions among participants (agents). In this framework, agents begin in a consistent state but diverge in their initial answers through spontaneous adversarial generation of Language Models (LLMs). Subsequently, the agents engage in debate to reach a consensus. During the same period, the MAD (Multi-Agent Debate) framework \citep{mad} introduces an asymmetric mechanism design. It assigns different roles (debater and judge) asymmetrically and utilizes different prompt decorators for similar roles (affirmative and negative sides). Other similar works include \cite{chan2023chateval}. Also, the ReConcile framework \citep{reconcile} exemplifies an asymmetric discussion mechanism by involving different LLMs and using distinct prompt decorators with varying convincing samples for each agent. To understand discussion more deeply, \cite{expd} try to explain such collaboration mechanism in a social psychology view. 

\noindent\textbf{Exploration Gaps: Discussion Frameworks, Prompt Decorators, and Mechanism Design.} Most of the existing works in this field primarily focus on evaluating the effectiveness of their own designed discussion frameworks or merely discussing the impact of different settings on the performance of these frameworks. However, there is a limited exploration of the potential effects that rich prompt decorator designs may have on the discussion process. Furthermore, there is currently a lack of unified definitions and discussions on Discussion Engineering and mechanism design based on language models. Additionally, little research has been conducted on examining the symmetry of mechanisms within discussions. Overall, more research is needed to understand the potential effects of rich prompt decorators and definition of mechanism design and symmetry.

\section{Discussion Engineering and Agent Symmetry} \label{sec: 3sym}

This section introduces a general framework to understand the discussion engineering of large language models. The discussion engineering, characterized by the mechanism $\mech=(M, \{(\llm_i, p_i)\}_{i=1}^n)$, includes careful designs at three levels: (1) the ($i$-th) inference $\llm_i$ of an LLM, (2) the prompt decorator $p_i$ for the inference $\llm_i$, and (3) the mechanism $M(\cdot, \{\llm_i, p_i\})$ to organize the inferences of LLMs. We further introduce the symmetry of the multi-agent system of LLMs to demonstrate the complexity of $\mech$. Under our framework, several examples are discussed.

\subsection{Agent symmetry in discussion engineering}
Symmetry and its breaking is a fundamental concept and widely investigated in science~\citep{anderson1972more}. Symmetry also plays an important role in the context of mechanism design of multiple agents~\citep{laffont2000mechanism,daskalakis2012symmetries}. Here we introduce a formal description to justify the agent symmetry of the mechanism $\mech$.

\begin{definition}[Computational graph]
Given the mechanism $\mech = (M, \{(\llm_i, p_i)\}_{i=1}^n)$, let $G = (V\cup\{x,y\}, E)$ be the directed graph whose node set $V\cup \{x,y\}$ includes the inference operation of LLM and the computational input node and output nodes. $x$ is the node to emit the input text, $y$ is the node that takes the inputs from one or multiple inferences and emits the final output, and $v_i\in V$ is the \textit{inference node} that describes the $i$-th  LLM inference with $(\llm_i, p_i)$. The directed edge in $(v_i, v_k)\in E$ describes the output of inference node $v_i$ is then the input of inference node $v_k$.
\end{definition}
We note that the graph $G$ contains all the necessary information to determine the mechanism $\mech$.

Then we introduce how the LLM agents are associated with the computational graph.
To make an LLM conversational session, the entire conversational history is always concatenated as part of the input of each inference call, resulting in many additional connections of the computational graph. The presence of agents allows us to conceptually assume the agents ``know'' the information in the conversation history, and then the complexity of the 
graph can be largely reduced. In this paper, we consider the computational graph with agents, so the connections that feed conversation history to the inference nodes are ignored for simplicity.

Let $\mathcal{A} = \{A_1, ..., A_m\}$ be the set of $m$ discussion agents. The inference call of an agent also concatenates its conversation history, denoted as $A_j(x)$.

\begin{definition}[Agent assignment]
Let $\mathcal{A} = \{A_1, ..., A_m\}$ be the set of $m$ discussion agents. Each inference node $v_i \in V$ is assigned to an agent $A_j\in \mathcal{A}$. Let $[n]$ present integers from $1$ to $n$, $\alpha: [n] \mapsto [m]$ is the assignment map that assigns the $i$-th inference to the $\alpha(i)$-th agent. Let $P\in \{0, 1\}^{n\times m}$ be the agent-inference assignment matrix, such that 
\begin{align}
    P_{ij} = \left\{\begin{array}{cc}
        1 & j = \alpha(i), \\
        0 & j \neq \alpha(i) 
    \end{array}\right. .
\end{align}
The assignment matrix $P$ is equivalent to the assignment map $\alpha$.
\end{definition}

Then a multi-agent discussion mechanism is defined by the triple $D = (\mech, \mathcal{A}, \alpha) = (G, \mathcal{A}, \alpha)$, which is the central object of discussion engineering.
Then, we can further introduce the concept of the coloring of the computational graph.
\begin{definition}[Agent coloring of the computational graph]
    Given the the multi-agent discussion $D = (G, \mathcal{A}, \alpha)$, then each inference node $v_i$ is described by $(\llm_i, p_i) = (A_{\alpha(i)}, p_i)$. Let $c_i^D = (A_{\alpha(i)}, p_i)$ be the color of $v_i$ and $C^D(v_i) = c_i^D$ be the color mapping of nodes, and $\mathcal{C}^D = (V, E, C^D)$ be a colored computational graph.
\end{definition}

Then the agent symmetry is established by the permutation operation over the agent set $\mathcal{A}$. Then we can discuss the symmetry by the agent permutation. We begin with the concept of mechanism invariance.
\begin{definition}[Mechanism invariance under the agent permutation]
    Given the multi-agent discussion $D = (G, \mathcal{A}, \alpha)$ and a permutation mapping $\pi: [m]\mapsto [m]$, the new discussion $D_\pi = (G, \mathcal{A}, \pi\circ\alpha)$ is derived by applying $\pi$ to the agents.
    We say a discussion $D$ is invariant under $\pi$ if and only if there is an isomorphism $\phi$ between two colored graphs $\mathcal{C}^D = (V, E, C^D)$ and $\mathcal{C}^{D_\pi} = (V, E, C^{D_\pi})$, such that
    \begin{enumerate}
        \item[(1)] $\phi$ is a bijection,
        \item[(2)] $\forall (v_i, v_j) \in E, (\phi(v_i), \phi(v_j)) \in E$,
        \item[(3)] $\forall v\in V, C^D(v) = C^{D_\pi}(\phi(v))$.
    \end{enumerate}
\end{definition}

We can also define the model invariance to justify the symmetry in the multi-model setting~\citep{reconcile}.
\begin{definition}[Model invariance under the agent permutation]
    Given the agent $\mathcal{A} = \{A_1, ..., A_m\}$, and their underlying LLM $\{L_1, .., L_m\}$ (i.e., ChatGPT, GPT4, etc.), the permutation $\pi: [m] \mapsto [m]$ is invariant if for $i=1,...,m$ $L_i = L_{\pi(i)}$.
\end{definition}

Following the definition of invariance, the symmetry group of agents is naturally defined following the standard algebra~\citep{tapp2021symmetry}. The largest possible symmetry group for a discussion of $m$ agents is the group $S_m$. However, the group $S_m$ is not always the symmetry group of a given multi-agent discussion mechanism $D$, the reasons for not achieving the largest symmetry group can be due to the asymmetry in the mechanism and \textbf{asymmetry in models}, respectively. Furthermore, for asymmetric mechanisms, there are two major ways of symmetry breaking. Referring to the three conditions of establishing the isomorphism, the dissatisfaction of condition (2) implies the asymmetry in the computational graph or \textbf{asymmetry in the mechanism pipeline} $M$, and that of condition (3) implies the \textbf{asymmetry in prompt decorators}.


\subsection{Examples of symmetries in discussion engineering} \label{sec:3.2}
We discuss the symmetry of several popular multi-agent discussion mechanisms
For simplicity, our goal is not to exhaustively enumerate the symmetry groups of mechanism symmetry or model symmetry.

\noindent\textbf{CoT-SC~\citep{cot-sc}.} The default setting of CoT-SC of $m$ agents reveals the largest symmetry group of $S_m$ in terms of mechanism invariance and model invariance.

\noindent\textbf{Debate~\citep{debate}.} The setting of $m$ agent debate also reveals $S_m$ in terms of mechanism invariance and model invariance

\noindent\textbf{MAD~\citep{mad}.} The largest possible mechanism symmetry group in MAD is $S_1$, where different agents play different roles. The model symmetry is $S_3$

\noindent\textbf{ReConcile~\citep{reconcile}.} The round-table discussion is an $S_3$ mechanism. However, the model is only $S_1$ for cases when the mechanism involves different LLMs.

\noindent\textbf{CR~\citep{cr}.} The cumulative reasoning also introduces agents with different roles, which is $S_3$.

\section{CMD: Conquer and Merge Discussion Framework}
\subsection{Motivation}
\begin{figure}[t]
    \centering
    \includegraphics[width=\textwidth]{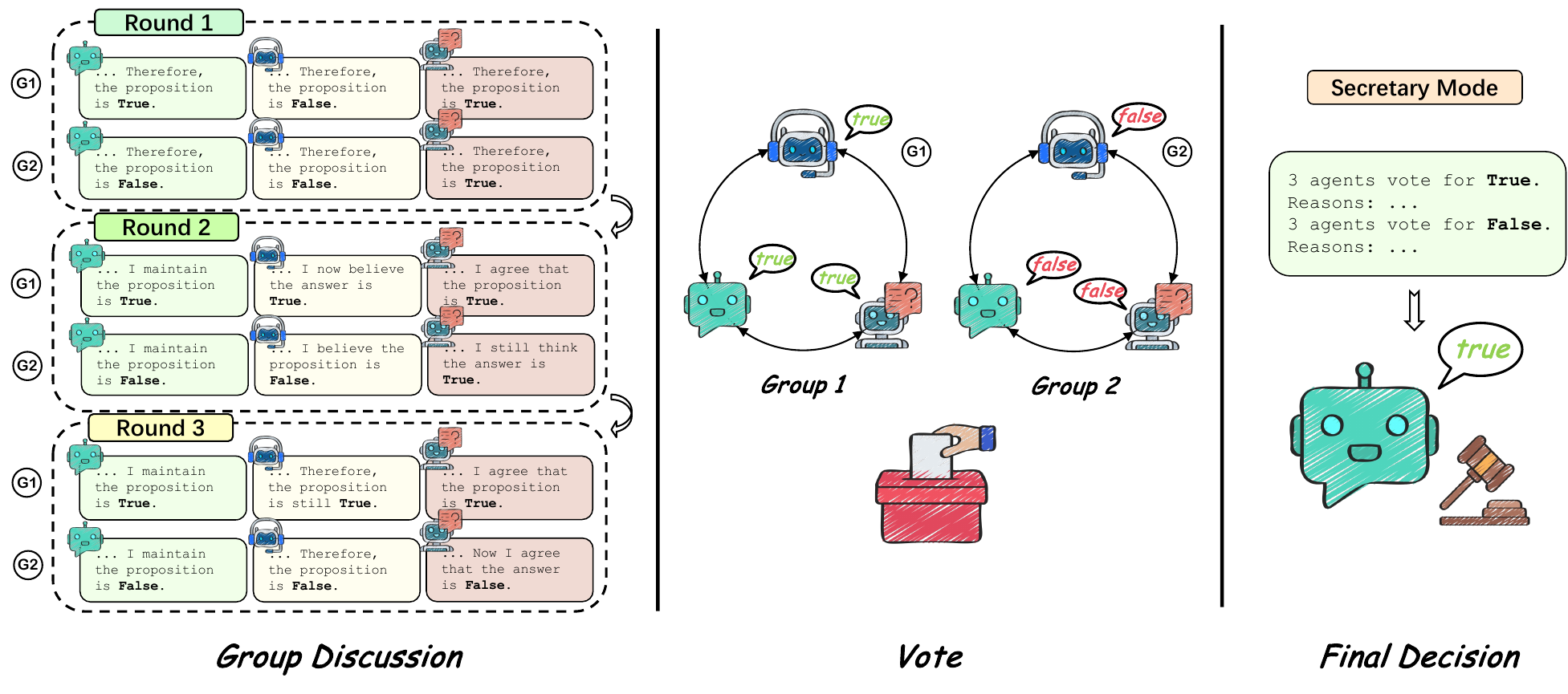}
    \caption{Overview of the Conquer-and-Merge Discussion (CMD) Framework.}
    \label{fig:cmd}
\end{figure}

The Debate framework proposes that an increased number of agents and discussion rounds will result in improved performance in multi-agent discussions \citep{debate}. Also, the ReConcile framework asserts that a greater number of discussion rounds leads to a higher level of consensus among agents, and the higher the consensus among agents, the more accurate the discussion outcomes become \citep{reconcile}. 
However, as the number of agents increases in a discussion, it leads to increased overhead in each round of discussion: 1. Each agent has to read more viewpoints from others, resulting in a sharp rise in input tokens. 2. The increase in input tokens puts significant pressure on language models with context token limitations. Inspired by real-life group discussions, we propose a variant framework for Debate called CMD. Fig \ref{fig:cmd} shows the overview of our method.\par

\subsection{Problem Definition}
Assume that there are $n$ agents $\mathcal{A}=\{A_i\}_{i=1}^m$ are discussing the given debate task $\mathcal{Q}$, and each agent is a session created from a LLM. Suppose that the maximum number of the discussion rounds is $R$, the current round is $r$, the current discussion level is $L$, and the current active agents set is $\mathcal{A'}$. For each agent $A_i\in\mathcal{A'}$, an answer it generates is $Ans_i=(v_i, e_i)$, where $v_i$ denotes the viewpoint and $e_i$ denotes the explanation. A debate history is $H^{(r)}=\bigcup_{i=1}^t Ans_i$ where $t=\left|\mathcal{A'}\right|$. $Ans_i^{(r)}$ is generated through the input prompt $A_i(\mathcal{Q}, Ans_i^{(r-1)}, O_i^{(r-1)})$, where $O_i^{(r)}$ indicates the opinions generated by $A_{j\neq i}$ in $(r-1)$-th round. The formal definition of $O_i^{(r)}$ is
\begin{equation}
    O_i^{(r)} = \bigcup_{j=1,j\neq i}^{\left|H^{(r-1)}\right|}\left(v_i, e_i\cdot\mathbbmss{1} \{\mathtt{Group}^{(L)}(A_i)=\mathtt{Group}^{(L)}(A_j)\} \right).
\end{equation}
This represents that $A_i$ will receive all information from the group members while it can only receive viewpoints from agents that are not in the same group. Our goal is to obtain the final result $a$ through an unweighted vote. In the event of a tie, either $a$ can be made by a secretary $S$, or representatives from each group will proceed to the next level of discussion until the tie is resolved. Each discussion group has only one representative. Note that $\mathcal{Q}$ can be described differently for each $A_i$. For example, different agents may be asked to hold different views at first.

\begin{algorithm}[t]
\caption{CMD: A Conquer and Merge Style Multi-Agents Discussion Framework}
\label{alg:CMD}
\begin{algorithmic}[1]
\Require Debate Task $\mathcal{Q}$, Maximum Discussion Rounds $R$, Agents $\mathcal{A}=\{A_i\}_{i=1}^n$, Secretary $S$
\Ensure Final Result $a$
\Function {CMD}{$\mathcal{Q},R,\mathcal{A},S$}
    \State $\mathcal{A'} \gets \mathcal{A}$ \Comment{Initialize current active agents}
    \State $N{g} \gets n / 3$ \Comment{Initialize group numbers, every 3 agents a group}
    \State $r \gets 0, L \gets 0$ \Comment{Initialize current round and current discussion level}
    \State $L_{max} \gets \mathtt{getMaxLevel}(\mathcal{A}, S)$
    \State $M_{g} \gets \mathtt{genGroupMap}(\mathcal{A}, N_{g}, L_{max})$ \Comment{Assign agents as groups based on discussion levels}
    \State $Ans_i\gets \emptyset, O_i \gets \emptyset, H\gets[]$ \Comment{Initialize answer, others' opinions and history}
    \While{$L \leq L_{max}$}
        \While{$r \leq R$}
            \If{$r>0$} \Comment{Stage 2}
                \State Update $O_i$ from $H$
            \EndIf
            \State $H\gets[]$
            \For{\textbf{each} $A_i \in \mathcal{A'}$}
                \State $Ans_i \gets A_i(\mathcal{Q}, Ans_i, O_i)$ \Comment{$Ans_i=(v_i, e_i)$}
                \State $H\gets H + [Ans_i]$
            \EndFor
            \State $r\gets r+1$
        \EndWhile
        \State $a \gets \mathtt{AnswerVote}(H)$
        \If{$a \neq $ \textbf{Tie}} \textbf{break}
        \ElsIf{$S$ \textbf{is not None}} \Comment{Check if the secretary mode is on}
            \State $a \gets S(\mathcal{Q}, H)$; \textbf{break}
        \Else \Comment{Secretary mode is off, representative mode is on}
            \State $L\gets L+1$
            \State $\mathcal{A'} \gets \mathtt{currentActiveAgents}(L, M_g)$
        \EndIf
        \State $r \gets 0$ \Comment{The higher-level discussion has commenced}
    \EndWhile
    \State \Return $a$
\EndFunction
\end{algorithmic}
\end{algorithm}

\subsection{CMD Stages}
Generally, there are three stages in \textbf{CMD}. In stage 1, the group map and all states will be initialized, then all the participants will generate their initial answers. Then, in stage 2, the participants will continue the discussion during the remaining rounds in groups. When the discussion round reaches the maximum number, it moves to stage 3. In this stage, all participants will vote to get the final answer.
\begin{itemize}[itemsep=5pt,topsep=0pt,parsep=0pt, leftmargin=12pt]
    \item $\mathsf{Stage\ 1:\ Group\ Discussion\ Initialization.}$ In this stage, $\mathcal{A'}$ will be initialized as all participants. All participating agents are initially assigned names in uppercase letters, and they are then grouped in sets of three. For $\mathtt{getMaxLevel}$, if secretary mode is on, then it will be set to $1$. Otherwise, it will be determined based on the current number of agents. For $\mathtt{getGroupMap}$, the algorithm will automatically generate the groups for all levels. Higher-level groups are generated among the current representatives. All active agents will generate their first response $Ans_i$.
    \item $\mathsf{Stage\ 2:\ Multi\ Rounds\ Discussion.}$ In this stage, the active agents will continue to discuss in the remaining rounds. Assume current round number is $r$, for each agent $A_i\in\mathcal{A'}$, 
    \begin{itemize}[itemsep=5pt,topsep=0pt,parsep=0pt, leftmargin=12pt]
        \item[1.] The algorithm will start to update the opinions history $O_i^{(r-1)}$ from $H$. For each record $h_j$ whose sender $A_j (j\neq i)$ stored in $H$, if $A_j$ and $A_i$ are in the same group, $O_i^{(r-1)}$ will record both the viewpoint and explanation from $h_j$; otherwise, $O_i^{(r-1)}$ will only record the viewpoint from $h_j$. After traversing all the records in $H$, $O_i$ will first gather all opinions of agents from other groups, then $O_i^{(r-1)}$ will gather local group explanations based on group members opinions.
        \item[2.] To save up tokens, last round history $H$ will be reset. Then, the new prompt $p_i^{(r)}$ will be generated based on $(\mathcal{Q}, Ans_i^{(r-1)}, O_i^{(r-1)})$. $A_i$ will make a response $Ans_i=(v_i,e_i)$ when given $p_i^{(r)}$. At last, $Ans_i$ will be appended to current round history $H$ for the further use.
        \item[3.] Repeat 1. and 2. until the maximum number of discussion rounds is reached.
    \end{itemize}
    \item $\mathsf{Stage\ 3:\ Vote\ for\ the\ Final\ Result.}$ In this stage, all agents will vote based on their final viewpoints stored in the history $H^{(r)}$ to obtain the result $a$. If there is no tie, then $a$ will be the final result, thus the discussion is over. If there is a tie and secretary mode is on, then the final result will be obtained through $S$ based on the viewpoints and explanations from all sides. If there is a tie and representative mode is on, then the discussion will move to the higher level, and deactivate agents that do not represent their groups. The representatives will be assigned to new groups, and return to $\mathsf{Stage\ 2}$ for further discussion. If there is still a tie, new representatives will be designated for further discussion, until either the tie is resolved or only one agent is activated for the discussion.
\end{itemize}

\section{Empirical Results on FOLIO-Wiki-Curated}
The synergy between prompt engineering and discussion engineering and its impact on the model's performance in downstream tasks has not been previously defined or studied. To better delineate the impact of the designed complexities in prompt engineering and discussion engineering, we provide a systematic evaluation on various settings that involve both prompt decorator designs and discussion engineering according to the symmetry mentioned in section \ref{sec: 3sym}.

\subsection{Implementation Details, Tasks and Metrics}

Considering the limitations of time and cost, all our experiments currently employ ChatGPT as the tested LLM, and all experiments are conducted using the \textit{FOLIO Wiki curated} \citep{folio, cr}. For each agent $A_i$, the language model $\llm$ it uses is \textit{GPT-3.5-Turbo-0613} from Azure OpenAI \footnote{\url{https://oai.azure.com/}}. We set the temperature to 0.25 for chat completion of ChatGPT.

We evaluate all configurations using the FOLIO wiki curated task, which is derived from the FOLIO dataset \citep{folio}. FOLIO is an open-domain dataset designed for natural language reasoning with first-order logic, encompassing a wide range of logically complex topics. The FOLIO wiki curated dataset consists of 460 cases sourced from wiki \footnote{\url{https://www.wikipedia.org/}}, and it has been carefully curated by \cite{cr} to remove any incorrect or ambiguous data cases. In our analysis, we report the accuracy of this task across all experimental settings.

\subsection{Single Agent - Prompt Decorator Designs}
Inspired by In-Context Learning \citep{gpt3}, CoT \citep{cot} and Natural Program \cite{np}, we devise a strategy to create prompts with different components, specifically tailored to the characteristics of downstream tasks. In Figure \ref{fig:pcomponets}, we present the schema outlining the design of prompt decorators for individual agents. Now we will provide a comprehensive overview of the four dimensions that we take into account while constructing these prompts:

\begin{figure}[t]
    \centering
    \includegraphics[width=\textwidth]{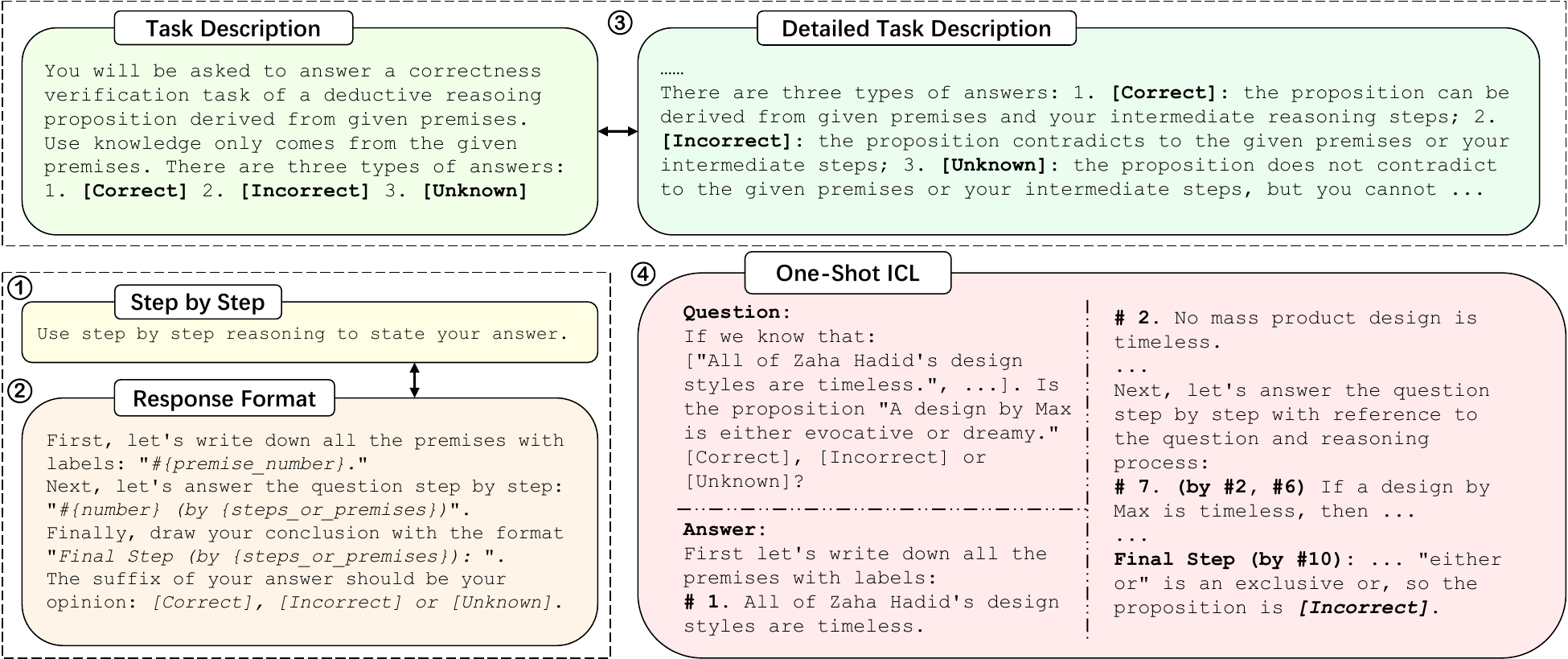}
    \caption{An illustration for different components of prompt decorators' designs. The dashed boxes in the figure indicate that the prompt components within the boxes are mutually interchangeable.}
    \label{fig:pcomponets}
\end{figure}

\begin{itemize}[itemsep=5pt,topsep=0pt,parsep=0pt, leftmargin=20pt]
    \item \textbf{Step by Step.} Step by step reasoning is commonly used to help the language model solve the given tasks by breaking down solutions into multiple intermediate steps. In our work, we have two ways to explicitly instruct the language model to reason step by step shown in figure \ref{fig:pcomponets} as \ding{172} and \ding{173}, one is to directly ask the agent to reason step by step, the other is implied within the required answer format.
    \item \textbf{Detailed Task Description.} Detailed task description includes two dimensions: 1. A general background and requirement introduction over the aiming task. 2. A detailed explanation over the possible answers to the aiming task. Given that there are three answer types of FOLIO wiki curated dataset: true, false and unknown, chances are that LLMs will find it confusing to distinguish false answer and unknown answer. The example is \ding{174} in figure \ref{fig:pcomponets}.
    \item \textbf{Response Format.} Response format refers to the required manner in which an agent should answer, following a specified format for responding. The example is \ding{173} in figure \ref{fig:pcomponets}.
    \item \textbf{One-Shot ICL.} One-Shot In-Context Learning indicates that one response example (possibly a $\langle\mathtt{question, answer}\rangle$ pair) is provided to help the agent understand the task definition and how it should respond. We regard it as an extension of the response format. The example is \ding{175} in figure \ref{fig:pcomponets}.
\end{itemize}

To better investigate the impact of varying complexity levels on agent performance, we conduct experiments using a single agent on the FOLIO wiki curated dataset. The obtained results are presented in Table \ref{tab:sape}.

\begin{table}[t]
\centering
\setlength\extrarowheight{4pt} 
\begin{tabularx}{\textwidth}{cccccc}
\Xhline{1.2pt} 
\thead{ID} & \thead{Step by Step} & \thead{Task Description} & \thead{Response Format} & \thead{One-Shot} & \thead{Acc (\%)} \\
\hline
$\mathtt{1}$ & $\boxwrong$ & $\boxwrong$ & $\boxwrong$ & $\boxwrong$ & 70.87\\
$\mathtt{2}$ & $\boxcheck$ & $\boxwrong$ & $\boxwrong$ & $\boxwrong$ & 70.22\\
$\mathtt{3}$ & $\boxcheck$ & $\boxcheck$ & $\boxwrong$ & $\boxwrong$ & 73.26\\
$\mathtt{4}$ & $\boxcheck$ & $\boxwrong$ & $\boxcheck$ & $\boxwrong$ & 71.30\\
$\mathtt{5}$ & $\boxcheck$ & $\boxcheck$ & $\boxcheck$ & $\boxwrong$ & 73.91\\
$\mathtt{6}$ & $\boxcheck$ & $\boxcheck$ & $\boxcheck$ & $\boxcheck$ & \textbf{76.09}\\[2pt]
\Xhline{1.2pt} 
\end{tabularx}
\caption{Single agent performance on FOLIO wiki curated with different level of complexity. Test with \textit{gpt-35-turbo-0613}.}
\label{tab:sape}
\end{table}

\noindent \textbf{Only prompt the "think step by step" will not improve the performance.} From the ID $\mathtt{1}$ and $\mathtt{2}$ in Table \ref{tab:sape}, we can learn that adding a simple sentence "please think step by step" into the prompt will not necessarily improve the performance, this is may be that the agent sometimes can automatically reason the proposition step by step without giving direct commands.

\noindent \textbf{Detailed task description will reduce answer ambiguity thus contributing to better performance.} From the ID $\mathtt{3}$ and $\mathtt{4}$ in Table \ref{tab:sape}, we can see the performance of detailed task description in prompt is better than that of response format in prompt. This result indicates that clarification for potentially ambiguous answer types will help LLMs improve performance on some tasks.

\noindent \textbf{One shot in-context learning can help the single agent better understand the question.} From the ID $\mathtt{5}$ and $\mathtt{6}$, we can clearly see that the performance when one shot example is introduced into the prompt is better than that of the zero shot settings. This result demonstrates that the in context learning is an essential contribution to the LLM performance, which also aligns with the previous works.

\subsection{Multi-Agent Discussion - Prompt and Discussion Engineering}
From section \ref{sec: 3sym}, we have defined the concept of discussion engineering symmetry. In this part, we will demonstrate the effects of different discussion frameworks with various symmetry settings and prompt decorators.

\subsubsection{Experiments Setup of Discussion Engineering }
In section \ref{sec: 3sym}, we categorized multi-agent discussions into symmetric and asymmetric discussions. Furthermore, we defined specific symmetry attributes in discussion engineering from three perspectives: (1) Prompt; (2) Agent; (3) Algorithm. Specific discussion methods are listed in section \ref{sec:3.2}. Furthermore, to better examine the impact of initial viewpoints on the convergence of the final answer in discussions, we introduced a prompt requirement for each agent assuming the role of a discussant. They are asked to hold a specific answer, creating prompt asymmetry. This ``hold different views'' setting only applies to the initial prompt and doesn't impose additional answer requirements during the subsequent discussion.
\noindent \textbf{}
\begin{table}[t]
\centering
\renewcommand{\arraystretch}{1.2}
\begin{tabular}{ccc}
\Xhline{1.2pt}
\textbf{Method}          & \textbf{Hold Different Views} & \textbf{Acc (\%)} \\[3pt] \hline
MAD                      & -                    & 74.13    \\
Debate (3 agents)        & Yes                  & 71.09    \\
Debate (6 agents)        & Yes                  & 73.04    \\
CMD (6 agents) & Yes                  & 76.30    \\
Debate (3 agents)        & No                   & 75.65    \\
Debate (6 agents)        & No                   & 74.38    \\
CMD (6 agents) & No                   & \textbf{77.39}    \\
\Xhline{1.2pt}
\end{tabular}
\caption{Performance on FOLIO wiki curated under different discussion settings using the \textit{gpt-35-turbo-0613} model. The "\textbf{Hold Different Views}" setting refers to asking agents to hold a specific opinion when generating the initial response. However, the MAD framework naturally incorporates different views, so the "Hold Different Views" option is not applicable for it.}
\label{tab:de}
\end{table}

\noindent\textbf{CMD Performs Better than Other Tested Methods.} The analysis of the results in table \ref{tab:de} reveals interesting insights. The CMD (6 agents) setting without Hold Different Views emerged as the most successful in terms of accuracy. Also, CMD achieves the best performance among tested discussion methods both with and without holding different views. This suggests that providing the viewpoints of all participants in the discussion along with partial explanations performs better than providing the viewpoints of all participants and their full explanations.

\noindent\textbf{Holding Different Views may Introduce Bias for Discussions.} 

On the other hand, the settings that introduced prompt asymmetry by requiring agents to hold different views during the initial prompt exhibits slightly lower accuracies. This implies that when agents begin with divergent perspectives, it may introduce more complexity and challenges to the discussion process, resulting in a slightly lower level of consensus in the final answer.


As figure \ref{tab:de} shows, it is worth noting that the differences in accuracy between the various settings are relatively small. While the setting without Hold Different Views performed slightly better, the overall impact of prompt asymmetry on the convergence of the final answer is minor. 

\subsubsection{The Entanglement between Prompt and Discussion Engineering}
Expanding on the design principles of the earlier prompt decorators, we extract the task-specific elements from those designs and incorporate them into different discussion settings for experimental purposes. Results can be seen in table \ref{tab:pe&de}.

\noindent\textbf{Discussion Mechanism Performance on Various Prompt Decorators.} Based on our observations of the experimental results, discussions consistently perform best when there is a Detailed Task Description or when all features are included. From table \ref{tab:pe&de}, we find that in all tested discussion mechanisms, the performance of the discussion initially improves and then declines as the features increase, only to improve again when all features are added. Our interpretation of this finding is that for FOLIO Wiki curated, language models sometimes struggle to distinguish between the relationships of true and unknown, and false and unknown. By providing detailed explanations for each answer type, we help the language model better understand the conclusions reached through reasoning. Additionally, having examples of answer formats that are more aligned with the task enhances the effectiveness of the discussion. Requesting a specific answer format without providing examples, on the other hand, introduces confusion during reasoning.

\noindent\textbf{Diminishing Marginal Returns of Complex Prompts and Discussions.} We observe that the prompt decorator that performs well on single-agent settings continues to be one of the best performers in discussions, but the accuracy decreases after the discussion, with only slight improvements observed in our proposed CMD method. For example, the all features added setting, the single agent achieves 76.09\% from table \ref{tab:sape}, while MAD, Debate (3 agents), Debate (6 agents) and CMD achieve 74.13\%, 75.65\%, 74.78\%, 77.39\% respectively from table \ref{tab:pe&de}. One possible reason for this phenomenon is that the combined effect of complex prompts and complex discussions leads to diminishing marginal returns, as both approaches are already approaching the limit of what the LLM can achieve when answering similar questions. Increasing the complexity of the mechanisms may introduce interference instead.

\noindent\textbf{Simple Prompts with Discussion Engineering 
Perform Better, and CMD Performs the Best.} Finally, from figure \ref{tab:pe&de}, we observe in the results that our proposed CMD method outperforms single-agent approaches in all cases except for zero-shot scenarios with specified answer formats. Additionally, under other discussion mechanisms, CMD performs better than the corresponding single agent when a Detailed Task Description is provided. This result indicates that when dealing with a task that is not particularly familiar, it is beneficial to first design a simple prompt decorator and then apply it to CMD. This approach allows for achieving good results without requiring a deep understanding of the task.

\begin{table}[t]
\centering
\setlength\extrarowheight{3pt}
\begin{tabular}{cccccc}
\Xhline{1.2pt}
\thead{Method} & \thead{Step by Step} & \thead{Task \\ Description} & \thead{Response \\Format} & \thead{One-Shot} & \thead{Acc (\%)} \\
\hline
\multirow{6}{*}{MAD} & $\boxwrong$ & $\boxwrong$ & $\boxwrong$ & $\boxwrong$ & 61.52\\
& $\boxcheck$ & $\boxwrong$ & $\boxwrong$ & $\boxwrong$ & 64.13\\
& $\boxcheck$ & $\boxcheck$ & $\boxwrong$ & $\boxwrong$ & 74.13\\
& $\boxcheck$ & $\boxwrong$ & $\boxcheck$ & $\boxwrong$ & 68.91\\
& $\boxcheck$ & $\boxcheck$ & $\boxcheck$ & $\boxwrong$ & 71.96\\
& $\boxcheck$ & $\boxcheck$ & $\boxcheck$ & $\boxcheck$ & 74.13\\
\hline
\multirow{5}{*}{Debate (3 agents)} 
& $\boxcheck$ & $\boxwrong$ & $\boxwrong$ & $\boxwrong$ & 70.00\\
& $\boxcheck$ & $\boxcheck$ & $\boxwrong$ & $\boxwrong$ & 75.65\\
& $\boxcheck$ & $\boxwrong$ & $\boxcheck$ & $\boxwrong$ & 71.96\\
& $\boxcheck$ & $\boxcheck$ & $\boxcheck$ & $\boxwrong$ & 70.22\\
& $\boxcheck$ & $\boxcheck$ & $\boxcheck$ & $\boxcheck$ & 75.65\\
\hline
\multirow{5}{*}{Debate (6 agents)} 
& $\boxcheck$ & $\boxwrong$ & $\boxwrong$ & $\boxwrong$ & 69.13\\
& $\boxcheck$ & $\boxcheck$ & $\boxwrong$ & $\boxwrong$ & 76.30\\
& $\boxcheck$ & $\boxwrong$ & $\boxcheck$ & $\boxwrong$ & 71.74\\
& $\boxcheck$ & $\boxcheck$ & $\boxcheck$ & $\boxwrong$ & 70.00\\
& $\boxcheck$ & $\boxcheck$ & $\boxcheck$ & $\boxcheck$ & 74.78\\
\hline
\multirow{4}{*}{CMD} & $\boxcheck$ & $\boxwrong$ & $\boxwrong$ & $\boxwrong$ & 73.26\\
& $\boxcheck$ & $\boxcheck$ & $\boxwrong$ & $\boxwrong$ & 74.13\\
& $\boxcheck$ & $\boxcheck$ & $\boxcheck$ & $\boxwrong$ & 71.09\\
& $\boxcheck$ & $\boxcheck$ & $\boxcheck$ & $\boxcheck$ & 77.39\\[2pt]
\Xhline{1.2pt} 
\end{tabular}
\caption{Performance on FOLIO wiki curated under different discussion settings using the \textit{gpt-35-turbo-0613} model. The table evaluates the accuracy of different discussion mechanisms (MAD, Debate, CMD) on the FOLIO wiki curated task using prompts with varying features.}
\label{tab:pe&de}
\end{table}


\subsubsection{Round Analysis among Discussions}
\begin{figure}[t]
    \centering
    \includegraphics[width=0.8\textwidth]{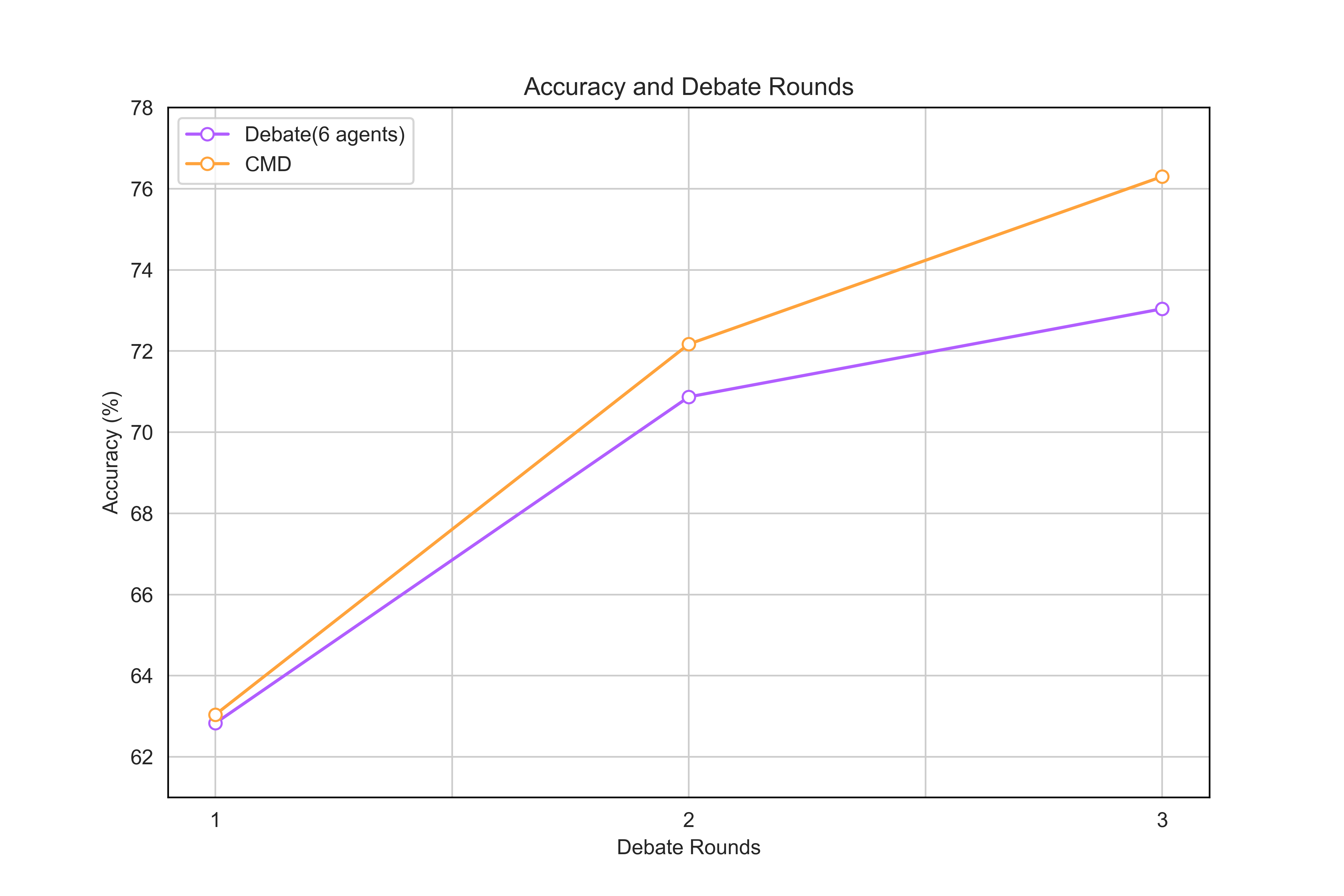}
    \caption{Accuracy of Debate (6 agents) and CMD at each round under the "Hold Different Views" setting. The graph shows how the accuracy of the Debate and CMD discussion mechanisms changes with the number of discussion rounds when agents are initially asked to hold different views.}
    \label{fig:rounds}
\end{figure}

To gain a deeper understanding of the dynamics within multi-agent discussions, we conducted further evaluations to assess the accuracy of each mechanism under the Hold Different Views setting. Specifically, we focused on evaluating the Debate (6 agents) and CMD mechanisms. The results of these evaluations are illustrated in Figure \ref{fig:rounds}.

Upon analyzing the figure, we observe a notable trend: as the number of discussion rounds increases, the accuracy steadily improves. This finding indicates that the discussion mechanism possesses the capability to rectify errors made by some agents as the debate progresses. By allowing for multiple rounds of debate and exchange of perspectives, the mechanism can leverage the collective intelligence of the agents to overcome individual biases and challenges. This iterative process enables the mechanism to refine and converge towards a more accurate and reliable final answer, ultimately improving the overall effectiveness of the multi-agent discussion.

\newpage
\section{Conclusion}

In this study, we explore various dimensions of combining prompt engineering and discussion mechanisms to improve reasoning abilities of language models. Our contributions can be summarized as follows:
\begin{itemize}[leftmargin=20pt]
    \item[(1)] We propose a theoretical framework to systematically characterize discussion engineering based on the concept of discussion mechanisms symmetry.
    \item[(2)] Empirically, our experiments reveal that both elaborate prompt designs and complex discussion mechanisms can approach the inherent performance limit of language models on a given task. Specifically, we show that reasoning capability of language models can be achieved by either a strong prompt but simple mechanism or a strong mechanism but a simple prompt.
    \item[(3)] We also propose a conquer and merge style (CMD) mechanism of group discussion, which is more scalable and outperforms other discussion mechanisms in many settings.
\end{itemize}

In summary, our work provides new insights on leveraging the synergy between prompt engineering and discussion engineering for LLMs. It highlights viable pathways to unlock reasoning capabilities with simple yet effective designs. Moving forward, more research can explore prompt and mechanism designs tailored to different model architectures and tasks. The theoretical frameworks established here also facilitate systematic exploration of new mechanism ideas.


\section*{Acknowledgement}

Zihao Wang thanks Mr. Datong Zhou from The Pennsylvania State University for the discussion on symmetry.

%% file: appendix.tex
\appendix
\section*{Appendix}
\section{Prompt Engineering} \label{app:A}
\subsection{Meta Prompt}

\begin{table}[H]
\centering
\setlength\extrarowheight{3pt}
\begin{tabularx}{\textwidth}{l}
\Xhline{1.2pt}
\textbf{System}\\[3pt]
\hline
You will be asked to answer a correctness verification task of a deductive reasoing proposition. \\
Use knowledge only comes from the given premises.
\\The suffix of your answer should be the your opinion of one of these answer type:
\\ 1. [Correct], 2. [Incorrect], 3. [Unknown].
\\Do not omit the brackets.\\[3pt]
\Xhline{1.2pt}
\end{tabularx}
\caption{Vanilla Prompt}
\label{tab:vanilla}
\end{table}

\begin{table}[H]
\centering
\setlength\extrarowheight{3pt}
\begin{tabularx}{\textwidth}{l}
\Xhline{1.2pt}
\textbf{System}\\[3pt]
\hline
You will be asked to answer a correctness verification task of a deductive reasoing proposition. \\
Use knowledge only comes from the given premises.
\\The suffix of your answer should be the your opinion of one of these answer type:
\\ 1. [Correct], 2. [Incorrect], 3. [Unknown].
\\Do not omit the brackets.
\\\textcolor{blue}{Use step-by-step reasoning to obtain your answer.}\\[3pt]
\Xhline{1.2pt}
\end{tabularx}
\caption{Step-by-Step Prompt}
\label{tab:cot}
\end{table}

\begin{table}[H]
\centering
\setlength\extrarowheight{3pt}
\begin{tabularx}{\textwidth}{X}
\Xhline{1.2pt}
\textbf{System}\\[3pt]
\hline
You will be asked to answer a correctness verification task of a deductive reasoning proposition. \\
Use knowledge only comes from the given premises.
\\Here are the instructions how you organize your answer format: \\
\textcolor{blue}{
First, let's write down all the premises with labels. The labels look like ``\#\{premise\_number\}''.
Next, let's answer the question step by step with reference to the question and reasoning process. There will be a prefix in your every reasoning step with the format ``\#\{number\} (by \{list\_of\_premises\_and\_steps\_used\})''.
}\\
\textcolor{blue}{In your final step, you should come to your conclusion with the format ``Final Step: (by \{list\_of\_premises\_and\_steps\_used\})''.}\\
The suffix of your answer should be the your opinion of one of these answer type:
\\ 1. [Correct], 2. [Incorrect], 3. [Unknown].
\\Do not omit the brackets.
\\[3pt]
\Xhline{1.2pt}
\end{tabularx}
\caption{Answer Format Prompt}
\label{tab:afp}
\end{table}

\begin{table}[H]
\centering
\setlength\extrarowheight{3pt}
\begin{tabularx}{\textwidth}{X}
\Xhline{1.2pt}
\textbf{System}\\[3pt]
\hline
You will be asked to answer a correctness verification task of a deductive reasoning proposition. \\
Use knowledge only comes from the given premises. There are three types of answers: \\
\textcolor{blue}{1. [Correct]: the proposition can be derived from given premises and your intermediate reasoning steps; 
2. [Incorrect]: the proposition contradicts the given premises or your intermediate steps; 
3. [Unknown]: the proposition does not contradict the given premises or your intermediate steps, but you cannot derive it from given premises and intermediate steps since there is lacking information in the given premises.}\\
The suffix of your answer should be the your opinion of one of these answer type:
\\ 1. [Correct], 2. [Incorrect], 3. [Unknown].Do not omit the brackets.
\\[3pt]
\Xhline{1.2pt}
\end{tabularx}
\caption{Detailed Task Description Prompt}
\label{tab:dtd}
\end{table}

\begin{table}[H]
\centering
\setlength\extrarowheight{3pt}
\begin{tabularx}{\textwidth}{X}
\Xhline{1.2pt}
\textbf{System}\\[3pt]
\hline
You will be asked to answer a correctness verification task of a deductive reasoning proposition. \\
Use knowledge only comes from the given premises.
\\Let me give you an example of a question and a well-formatted answer: \\
\textcolor{blue}{\bf Question:}\\
\textcolor{blue}{If we know that:
["All of Zaha Hadid's design styles are timeless.", "No mass product design is timeless.", "Either Zaha Hadid's design style or Kelly Wearstler's design style. ", "All of Kelly Wearstler's design styles are evocative.", "All of Kelly Wearstler's design styles are dreamy.", "If a design by Max is timeless, then a design by Max is a mass product design and evocative."],}\\
\textcolor{blue}{Is the proposition ""A design by Max is either evocative or dreamy." [Correct], [Incorrect] or [Unknown]?}\\
\textcolor{blue}{\bf Answer:}\\
\textcolor{blue}{First let's write down all the premises with labels:}\\
\textcolor{blue}{\# 1. All of Zaha Hadid's design styles are timeless.}\\
\textcolor{blue}{\# 2. No mass product design is timeless.}\\
\textcolor{blue}{\# 3. Either Zaha Hadid's design style or Kelly Wearstler's design style.}\\
\textcolor{blue}{\# 4. All of Kelly Wearstler's design styles are evocative.}\\
\textcolor{blue}{\# 5. All of Kelly Wearstler's design styles are dreamy.}\\
\textcolor{blue}{\# 6. If a design by Max is timeless, then a design by Max is a mass product design and evocative.}\\
\textcolor{blue}{Next, let's answer the question step by step with reference to the question and reasoning process:}\\
\textcolor{blue}{\# 7. (by \#2, \#6) If a design by Max is timeless, then a design by Max is a mass product design. However, \#2 states that no mass product design is timeless, so a design by Max is not timeless.}\\
\textcolor{blue}{\# 8. (by \#1, \#7) Since a design by Max is not timeless, \#1 states that a design by Max is not a design by Zaha Hadid.}\\
\textcolor{blue}{\# 9. (by \#3, \#8) Since a design by Max is not a design by Zaha Hadid, \#3 states that a design by Max is a design by Kelly Wearstler.}\\
\textcolor{blue}{\# 10. (by \#4, \#5, \#9) Since a design by Max is a design by Kelly Wearstler, \#4 states that a design by Max is evocative, \#5 states that a design by Max is dreamy.}\\
\textcolor{blue}{Final Step (by \#10): \#10 states that a design by Max is both evocative and dreamy. However, the proposition states that a design by Max is either evocative or dreamy. "either...or..." is an exclusive or, so the proposition is [Incorrect].}\\[3pt]
\Xhline{1.2pt}
\end{tabularx}
\caption{One Shot Prompt}
\label{tab:osp}
\end{table}

\subsection{An Answer Example with All Features Added}
\begin{table}[H]
\centering
\setlength\extrarowheight{3pt}
\begin{tabularx}{\textwidth}{X}
\Xhline{1.2pt}
\textbf{System}\\[3pt]
\hline
You will be asked to answer a correctness verification task of a deductive reasoning proposition. \\
Use knowledge only comes from the given premises.\\
$[$Detailed Task Description$]$ \\
$[$Answer Format$]$\\
Let me give you an example of a question and a well-formatted answer: \\
{\bf Question:}\\
{If we know that:
["All of Zaha Hadid's design styles are timeless.", "No mass product design is timeless.", "Either Zaha Hadid's design style or Kelly Wearstler's design style. ", "All of Kelly Wearstler's design styles are evocative.", "All of Kelly Wearstler's design styles are dreamy.", "If a design by Max is timeless, then a design by Max is a mass product design and evocative."],}\\
{Is the proposition ""A design by Max is either evocative or dreamy." [Correct], [Incorrect] or [Unknown]?}\\
{\bf Answer:}\\
{First let's write down all the premises with labels:}\\
{\# 1. All of Zaha Hadid's design styles are timeless.}\\
{\# 2. No mass product design is timeless.}\\
{\# 3. Either Zaha Hadid's design style or Kelly Wearstler's design style.}\\
{\# 4. All of Kelly Wearstler's design styles are evocative.}\\
{\# 5. All of Kelly Wearstler's design styles are dreamy.}\\
{\# 6. If a design by Max is timeless, then a design by Max is a mass product design and evocative.}\\
{Next, let's answer the question step by step with reference to the question and reasoning process:}\\
{\# 7. (by \#2, \#6) If a design by Max is timeless, then a design by Max is a mass product design. However, \#2 states that no mass product design is timeless, so a design by Max is not timeless.}\\
{\# 8. (by \#1, \#7) Since a design by Max is not timeless, \#1 states that a design by Max is not a design by Zaha Hadid.}\\
{\# 9. (by \#3, \#8) Since a design by Max is not a design by Zaha Hadid, \#3 states that a design by Max is a design by Kelly Wearstler.}\\
{\# 10. (by \#4, \#5, \#9) Since a design by Max is a design by Kelly Wearstler, \#4 states that a design by Max is evocative, \#5 states that a design by Max is dreamy.}\\
{Final Step (by \#10): \#10 states that a design by Max is both evocative and dreamy. However, the proposition states that a design by Max is either evocative or dreamy. "either...or..." is an exclusive or, so the proposition is [Incorrect].}\\[28pt]
\hline
\textbf{User}\\[3pt]
\hline
\textbf{Question:}\\
If we know that: [Neocrepidodera Corpulentas are flea beetles or moths. The species Neocrepidodera Corpulenta is in the Chrysomelidae family. There are no moths within the Chrysomelidae family.]\\
Is the proposition ``There are no flea beetles within the Chrysomelidae family.'' [Correct], [Incorrect] or [Unknown]?\\[16pt]
\Xhline{1.2pt}
\end{tabularx}
\caption{Question Prompt}
\label{tab:metaPrompt}
\end{table}

\begin{table}[H]
\centering
\setlength\extrarowheight{3pt}
\begin{tabularx}{\textwidth}{X}
\Xhline{1.2pt}
\textbf{Assistant (ChatGPT-3.5) - Single Agent}\\[3pt]
\hline
First let's write down all the premises with labels:\\
\#1. Neocrepidodera Corpulentas are flea beetles or moths.\\
\#2. The species Neocrepidodera Corpulenta is in the Chrysomelidae family.\\
\#3. There are no moths within the Chrysomelidae family.\\\\
Next, let's answer the question step by step with reference to the question and reasoning process:\\
\#4. (by \#1, \#2) Neocrepidodera Corpulenta is a species of Neocrepidodera Corpulentas, so it is either a flea beetle or a moth.\\
\#5. (by \#3, \#4) However, \#3 states that there are no moths within the Chrysomelidae family, so Neocrepidodera Corpulenta cannot be a moth.\\
Final Step (by \#5): Therefore, Neocrepidodera Corpulenta must be a flea beetle. The proposition ``There are no flea beetles within the Chrysomelidae family'' contradicts the given premises, so it is [Incorrect].\\[26pt]
\Xhline{1.2pt}
\end{tabularx}
\caption{ChatGPT Agent Answer}
\label{tab:agentAnswer}
\end{table}

\section{Discussion Engineering - CMD}
In the upcoming demonstration, we'll walk you through the CMD framework from the viewpoint of agent \textbf{A} among six agents. To keep things straightforward, we'll use the full-featured prompt decorator explained in Appendix \ref{app:A}. This choice is for convenience and to make the process easy to follow.
\subsection{Meta Prompt}
The meta prompt for agent \textbf{A} is the same as shown in the table \ref{tab:metaPrompt}.

\subsection{Round 1 Answer}

\begin{table}[H]
\centering
\setlength\extrarowheight{3pt}
\begin{tabularx}{\textwidth}{X}
\Xhline{1.2pt}
\textbf{Assistant (ChatGPT-3.5) - Agent A}\\[3pt]
\hline
First let's write down all the premises with labels:\\
\#1. Neocrepidodera Corpulentas are flea beetles or moths.\\
\#2. The species Neocrepidodera Corpulenta is in the Chrysomelidae family.\\
\#3. There are no moths within the Chrysomelidae family.\\\\
Next, let's answer the question step by step with reference to the question and reasoning process:\\
\#4. (by \#1, \#2) Neocrepidodera Corpulenta is either a flea beetle or a moth, and it is in the Chrysomelidae family.\\
\#5. (by \#3, \#4) Since there are no moths within the Chrysomelidae family, Neocrepidodera Corpulenta must be a flea beetle.\\
Final Step (by \#5): Neocrepidodera Corpulenta is a flea beetle, so the proposition ``There are no flea beetles within the Chrysomelidae family'' is [Incorrect].\\[15pt]
\Xhline{1.2pt}
\end{tabularx}
\caption{The Answer of Agent \textbf{A} in Round 1}
\label{tab:Aans1}
\end{table}

\subsection{Middle System and User Prompts in Round 1}

\begin{table}[H]
\centering
\setlength\extrarowheight{3pt}
\begin{tabularx}{\textwidth}{X}
\Xhline{1.2pt}
\textbf{System}\\[3pt]
\hline
There are 2 groups of people discussing on the same topic. I will provide you the detailed opinions and reasoning steps from your group member and opinions from other group members. Use these opinions and your previous opinion as additional advice, note that they maybe wrong. Do not copy other's entire answer, modify the part you believe is wrong.\\
Other group members' opinions:\\
\textcolor{blue}{Three agents think the proposition is Incorrect.}\\
Your group's opinions:\\
\textcolor{blue}{One agent thinks the proposition is Correct.} Below is his answer:\\
First let's write down all the premises with labels:\\
$\cdots$\\
Next, let's answer the question step by step with reference to the question and reasoning process:\\
\#4. (by \#1, \#2) Neocrepidodera Corpulenta is either a flea beetle or a moth, and it is in the Chrysomelidae family.\\
\#5. (by \#3) There are no moths within the Chrysomelidae family.\\
\#6. (by \#4, \#5) Neocrepidodera Corpulenta cannot be a moth because there are no moths in the Chrysomelidae family.\\
Final Step (by \#6): The proposition ``There are no flea beetles within the Chrysomelidae family'' is [Correct] because Neocrepidodera Corpulenta, which is in the Chrysomelidae family, cannot be a moth according to premise \#5.\\
\textcolor{blue}{One agent thinks the proposition is Incorrect.} Below is his answer:\\
$\cdots$\\
Next, let's answer the question step by step with reference to the question and reasoning process:\\
\#4. (by \#1, \#2) Neocrepidodera Corpulenta is a species of Neocrepidodera Corpulentas, so it is either a flea beetle or a moth.\\
\#5. (by \#3, \#4) Since there are no moths within the Chrysomelidae family, Neocrepidodera Corpulenta must be a flea beetle.\\
Final Step (by \#5): The proposition states that there are no flea beetles within the Chrysomelidae family. However, from premise \#2, we know that Neocrepidodera Corpulenta, which is a flea beetle, is in the Chrysomelidae family. Therefore, the proposition is [Incorrect].
\\[28pt]
\hline
\textbf{User}\\[3pt]
\hline
Use the provided opinions and your previous answer as additional advice critically, please provide an updated answer. Make sure to state your opinion over proposition at the end of the response in the given format: [Correct], [Incorrect] or [Unknown], and such format should be only used to determine the correctness of the proposition.\\[24pt]
\Xhline{1.2pt}
\end{tabularx}
\caption{Middle System Prompt at the End of Round 1}
\label{tab:midR1}
\end{table}

\subsection{Round 2 Answer}

\begin{table}[H]
\centering
\setlength\extrarowheight{3pt}
\begin{tabularx}{\textwidth}{X}
\Xhline{1.2pt}
\textbf{Assistant (ChatGPT-3.5) - Agent A}\\[3pt]
\hline
Based on the opinions provided by the other group members and my previous answer, I will critically evaluate the reasoning steps and provide an updated answer.\\
The first agent from my group argues that the proposition is correct. Their reasoning is as follows:\\
\#4. (by \#1, \#2) Neocrepidodera Corpulenta is either a flea beetle or a moth, and it is in the Chrysomelidae family.\\
\#5. (by \#3) There are no moths within the Chrysomelidae family.\\
\#6. (by \#4, \#5) Neocrepidodera Corpulenta cannot be a moth because there are no moths in the Chrysomelidae family.\\
Final Step (by \#6): The proposition ``There are no flea beetles within the Chrysomelidae family'' is [Correct] because Neocrepidodera Corpulenta, which is in the Chrysomelidae family, cannot be a moth according to premise \#5.\\
The second agent from my group argues that the proposition is incorrect. Their reasoning is as follows:\\
\#4. (by \#1, \#2) Neocrepidodera Corpulenta is a species of Neocrepidodera Corpulentas, so it is either a flea beetle or a moth.\\
\#5. (by \#3, \#4) Since there are no moths within the Chrysomelidae family, Neocrepidodera Corpulenta must be a flea beetle.\\
Final Step (by \#5): The proposition states that there are no flea beetles within the Chrysomelidae family. However, from premise \#2, we know that Neocrepidodera Corpulenta, which is a flea beetle, is in the Chrysomelidae family. Therefore, the proposition is [Incorrect].\\
After critically evaluating the reasoning steps, I agree with the second agent from my group that the proposition is incorrect. Neocrepidodera Corpulenta is a species of Neocrepidodera Corpulentas, which are either flea beetles or moths according to premise \#1. Since Neocrepidodera Corpulenta is in the Chrysomelidae family according to premise \#2, it cannot be concluded that there are no flea beetles within the Chrysomelidae family. Therefore, the proposition is [Incorrect].\\[50pt]
\Xhline{1.2pt}
\end{tabularx}
\caption{The Answer of Agent \textbf{A} in Round 2}
\label{tab:Aans2}
\end{table}

\subsection{Middle System Prompt at the End of Round 2}

\begin{table}[H]
\centering
\setlength\extrarowheight{3pt}
\begin{tabularx}{\textwidth}{X}
\Xhline{1.2pt}
\textbf{System}\\[3pt]
\hline
There are 2 groups of people discussing on the same topic. I will provide you the detailed opinions and reasoning steps from your group member and opinions from other group members. Use these opinions and your previous opinion as additional advice, note that they maybe wrong. Do not copy other's entire answer, modify the part you believe is wrong.\\
Other group members' opinions:\\
\textcolor{blue}{Three agents think the proposition is Incorrect.}\\
Your group's opinions:\\
\textcolor{blue}{Two agents think the proposition is Incorrect.} Below are their answers:\\
$\cdots$
\\
\hline
\textbf{User}\\[3pt]
\hline
This part is the same user part shown in table \ref{tab:midR1}.\\
\Xhline{1.2pt}
\end{tabularx}
\caption{Middle System Prompt at the End of Round 2}
\label{tab:midR2}
\end{table}

\subsection{Round 3 Answer}

\begin{table}[H]
\centering
\setlength\extrarowheight{3pt}
\begin{tabularx}{\textwidth}{X}
\Xhline{1.2pt}
\textbf{Assistant (ChatGPT-3.5) - Agent A}\\[3pt]
\hline
Based on the opinions provided by the other group members and my previous answer, here is an updated response:\\
First, let's write down all the premises with labels:\\
\#1. Neocrepidodera Corpulentas are flea beetles or moths.\\
\#2. The species Neocrepidodera Corpulenta is in the Chrysomelidae family.\\
\#3. There are no moths within the Chrysomelidae family.\\
Next, let's answer the question step by step with reference to the question and reasoning process:\\
\#4. (by \#1, \#2) Neocrepidodera Corpulenta is either a flea beetle or a moth, and it is in the Chrysomelidae family.\\
\#5. (by \#3) There are no moths within the Chrysomelidae family.\\
\#6. (by \#4, \#5) Neocrepidodera Corpulenta cannot be a moth because there are no moths in the Chrysomelidae family.\\
Final Step (by \#6): The proposition ``There are no flea beetles within the Chrysomelidae family'' contradicts the given premises, as Neocrepidodera Corpulenta, which is in the Chrysomelidae family, is a flea beetle. Therefore, the proposition is [Incorrect].\\
After critically evaluating the reasoning steps and considering the opinions provided by the other group members, I agree that the proposition is [Incorrect]. The given premises state that Neocrepidodera Corpulenta is in the Chrysomelidae family and that it is either a flea beetle or a moth. Since there are no moths within the Chrysomelidae family, it can be concluded that Neocrepidodera Corpulenta must be a flea beetle. Therefore, the proposition ``There are no flea beetles within the Chrysomelidae family'' contradicts the given premises.\\[64pt]
\Xhline{1.2pt}
\end{tabularx}
\caption{The Answer of Agent \textbf{A} in Round 3}
\label{tab:Aans3}
\end{table}

\section{CMD Secretary - A Tie Case Solution}
In the event of a tie between two groups, we suggest employing a secretary to resolve such cases. The complete chat transcript below in table \ref{tab:secretary} illustrates a scenario where the secretary provides the ultimate resolution.
\begin{table}[H]
\centering
\setlength\extrarowheight{3pt}
\begin{tabularx}{\textwidth}{X}
\Xhline{1.2pt}
\textbf{System}\\[3pt]
\hline
6 agents are discussing the correctness of a proposition based on the given premises. The given premises are: $\cdots$ The proposition is: ``IBM has an office in London or Zurich.''\\
However, now there is a draw:\\
Three agents think the proposition is Correct. Below is one of their answers: $\cdots$\\
Three agents think the proposition is Unknown. Below is one of their answers: $\cdots$\\
\hline
\textbf{User}\\[3pt]
\hline
Your task is to carefully determine which opinion is more plausible. Answer opinion types are: [Correct], [Incorrect], [Unknown]. You should give your response in the required format. You are forbidden to copy others' reasoning steps. You can only use the given premises and your reasoning steps to answer the question.\\
Here are the instructions how you organize your answer format:\\
First, let's write down all the premises with labels. The labels look like ``\#\{premise\_number\}.''\\
Next, let's answer the question step by step with reference to the question and reasoing process. There will be a prefix in your every reasoning step with the format ``\#\{number\} (by \{list\_of\_premises\_and\_steps\_used\})''. In your final step, you should come to your conclusion with the format ``Final Step (by \{list\_of\_premises\_and\_steps\_used\}):''.\\
The suffix of your answer should be the answer type: [Correct], [Incorrect] or [Unknown].\\
Here are several hints to help you conclude the final answer:\\
1. Pay attention to the logic of quantifiers and distinguish the meanings of quantifiers such as ``all'' and ``some''.\\
2. When two quantifiers are connected as the premise of existence, the conclusion may not be established, that is, it is unknown.
\\[16pt]
\hline
\textbf{Assistant (ChatGPT-3.5) - Secretary}\\[3pt]
\hline
First, let's write down all the premises with labels:\\
\#1. Evangelos Eleftheriou is a Greek electrical engineer.\\
\#2. Evangelos Eleftheriou worked for IBM in Zurich.\\
\#3. If a company has employees working for them somewhere, then they have an office there.\\
\#4. IBM is a company.\\
Next, let's answer the question step by step with reference to the question and reasoning process:\\
\#5 (by \#4, \#3) Since IBM is a company, and according to premise \#3, if a company has employees working for them somewhere, then they have an office there. Therefore, IBM has an office in Zurich.\\
Now, let's consider the opinions of the other group members:\\
- One agent thinks the proposition is Correct.\\
- Two agents think the proposition is Unknown.\\
Considering the opinions of the other group members, there is disagreement regarding the correctness of the proposition. However, based on the premises and the reasoning process, it can be concluded that IBM has an office in Zurich.\\
Final Step (by \#5): Since IBM has an office in Zurich, the proposition ``IBM has an office in London or Zurich'' is [Correct].\\[16pt]
\Xhline{1.2pt}
\end{tabularx}
\caption{Secretary to solve a tie case.}
\label{tab:secretary}
\end{table}